\documentclass[11pt]{article}

\usepackage[margin=1in]{geometry}
\usepackage[T1]{fontenc}
\usepackage[utf8]{inputenc}
\usepackage{lmodern}
\usepackage{booktabs}
\usepackage{tabularx}
\usepackage{placeins}
\usepackage{graphicx}
\usepackage{amsmath}
\usepackage{hyperref}
\usepackage{natbib}
\usepackage{enumitem}
\usepackage{xcolor}
\hypersetup{colorlinks=true,linkcolor=blue,citecolor=blue,urlcolor=blue}
\title{Uncertainty-Aware Last-Layer Adaptation of RETFound for Referable Diabetic Retinopathy Screening Under Dataset Shift}
\author{
Karim Mardhani \\
Master of Science in Artificial Intelligence \\
University of Colorado Boulder \\
Boulder, CO, USA \\
\texttt{Karim.Mardhani@colorado.edu}
}
\date{}

\begin{document}
\maketitle

\begin{center}
\small
Repository: \url{https://github.com/kmardhani/uncertainty-aware-retfound} \\
arXiv categories: \texttt{cs.CV} (primary), \texttt{cs.LG} (secondary)
\end{center}

\begin{abstract}
This paper presents a safety-centered empirical evaluation of uncertainty-aware last-layer adaptation for referable diabetic retinopathy screening using RETFound, a self-supervised vision-transformer retinal foundation model used here as a frozen feature encoder, and the public APTOS 2019 and DDR diabetic retinopathy fundus-image datasets. We compare a cached-feature softmax head, post-hoc temperature scaling, variational Bayesian last-layer heads, a diagonal Laplace last-layer approximation, and an SNGP-style cached-feature head. On APTOS, uncertainty-aware operating points improved sensitivity and selective-referral behavior. The strongest APTOS selective-referral result deferred approximately 20\% of cases and reduced accepted-case false negatives to zero while preserving high accepted-case specificity. However, threshold tuning also reduced false negatives at high false-positive cost, so false-negative reduction alone was not unique to Bayesian modeling. On DDR, native Bayesian heads qualitatively reproduced the APTOS direction but with weaker tradeoffs, while the APTOS-trained SNGP checkpoint transferred poorly and failed to provide useful external selective-referral behavior. These results highlight the value of safety-centered evaluation beyond aggregate accuracy: uncertainty-aware last-layer heads can improve internal safety-oriented operating points, but trustworthy retinal screening claims require explicit safety--coverage evaluation and second-dataset validation under shift.
\end{abstract}

\section{Introduction}
Deep retinal screening systems are often evaluated with aggregate discrimination metrics such as area under the ROC curve and overall accuracy. Those metrics matter, but they do not directly answer the safety question that matters most for screening: how many referable cases are missed, under which operating point, and with what human-review burden. In referable diabetic retinopathy (DR) screening, false negatives are especially consequential because a missed positive case can delay referral and treatment \citep{gulshan2016development,ting2017development}.

This project studies a deliberately limited setting: frozen RETFound retinal features \citep{zhou2023retfound} combined with small last-layer adaptation methods. The goal is not to claim a new foundation-model state of the art. The goal is to ask a narrower and more practical question: can uncertainty-aware last-layer heads produce safer operating points than a deterministic cached-feature baseline, and do those gains survive a second-dataset validation check?

The study evaluates five methodological families:
\begin{enumerate}[leftmargin=1.2em]
\item a cached-feature softmax linear head;
\item post-hoc temperature scaling \citep{guo2017calibration};
\item variational Bayesian last-layer heads;
\item a diagonal Laplace last-layer approximation \citep{ritter2018scalable,daxberger2021laplace};
\item an SNGP-style cached-feature head with random Fourier features and a diagonal precision proxy \citep{liu2020simple}.
\end{enumerate}

The main result is mixed, which is scientifically useful. On APTOS, uncertainty-aware last-layer heads improved high-sensitivity operating points and selective-referral behavior. On DDR, native Bayesian heads qualitatively reproduced the APTOS direction, but with weaker specificity and no zero-false-negative operating point at practical coverage. The APTOS-trained SNGP checkpoint evaluated directly on DDR without retraining performed poorly, providing a negative transfer result rather than evidence of shift robustness.

\paragraph{Main claim.}
Uncertainty-aware last-layer heads can improve internal safety-oriented operating points, but false-negative reduction is not unique to Bayesian modeling, threshold tuning can also reduce false negatives at high false-positive cost, and uncertainty estimates that appear useful internally may fail under external dataset shift. Therefore, uncertainty-aware retinal screening systems should be evaluated using explicit safety--coverage tradeoffs and second-dataset validation.

\section{Related Work}
RETFound is a retinal foundation model pretrained for ophthalmic image representation learning and has become a natural frozen-backbone starting point for downstream retinal tasks \citep{zhou2023retfound}. Our work uses RETFound as a fixed feature extractor rather than fine-tuning it end to end, which keeps the methodological question focused on last-layer uncertainty rather than representation learning.
Recent ophthalmic foundation-model studies have also begun to compare and combine eye-specific foundation models across multiple retinal and ophthalmic diagnostic tasks, including FusionFM, which evaluates several ophthalmic foundation models and explores fusion strategies for ophthalmic diagnosis~\citep{zou2025fusionfm}.

Uncertainty estimation for neural networks has a large literature, including approximate Bayesian inference, variational methods, Laplace approximations, and distance-aware posteriors \citep{neal1996bayesian,blundell2015weightuncertainty,ritter2018scalable,daxberger2021laplace,liu2020simple}. In medical imaging, calibration and uncertainty matter because ranking errors by confidence may support safer decision support or defer-to-human workflows. That motivation aligns with selective prediction and selective classification work, where a model is allowed to abstain on uncertain examples instead of making every decision automatically \citep{elyaniv2010foundation,geifman2017selective}.

What is still easy to overclaim in this area is causal uniqueness. If a Bayesian or uncertainty-aware model reduces false negatives, that does not by itself prove that Bayesian modeling was required. A lower decision threshold may sometimes do the same thing with worse false-positive burden. Our study therefore reports both threshold-sweep controls and selective-referral analyses, and it tests whether internal findings survive a second dataset.

\section{Materials and Methods}

\subsection{Datasets and binary referable-DR mapping}
We study binary referable DR classification on two datasets. APTOS 2019 is a public diabetic retinopathy grading dataset of color fundus photographs, while DDR is a larger public diabetic retinopathy grading dataset used here as a second-dataset validation setting. Native DDR softmax and Bayesian heads test whether the APTOS findings qualitatively reproduce on a harder second dataset. The APTOS-trained SNGP checkpoint evaluated directly on DDR is an APTOS-to-DDR transfer evaluation without retraining.

\begin{table}[t]
\centering
\small
\caption{Dataset sizes after binary referable diabetic retinopathy mapping.}
\label{tab:dataset-summary}
\begin{tabular}{lrrrrrr}
\toprule
Dataset & Total & Non-ref & Ref & Train & Val & Test \\
\midrule
APTOS 2019 & 3662 & 2175 & 1487 & 2930 & 366 & 366 \\
DDR & 12522 & 6896 & 5626 & 8765 & 1878 & 1879 \\
\bottomrule
\end{tabular}
\end{table}

For both datasets, grades 0 and 1 were mapped to non-referable DR, and grades 2, 3, and 4 were mapped to referable DR. That mapping is clinically motivated by the screening question of whether a case should be referred onward for further ophthalmic review.

\subsection{Frozen RETFound feature extraction}
All downstream models operate on frozen RETFound feature vectors rather than raw pixels. Features were extracted from the \texttt{RETFound\_mae} checkpoint \texttt{RETFound\_mae\_natureCFP.pth}. Each image is passed through the pretrained RETFound encoder, and the resulting 1024-dimensional feature vector is used to train lightweight last-layer classifiers. Images were resized and center-cropped to 224$\times$224, and the exported feature dimensionality was 1024 for both datasets.

\begin{table}[t]
\centering
\footnotesize
\caption{Frozen RETFound feature extraction settings used for cached-feature experiments.}
\label{tab:feature-summary}
\begin{tabularx}{\textwidth}{l l X r r r l r}
\toprule
Dataset & Backbone & Checkpoint & Dim & Resize & Crop & Device & Batch \\
\midrule
APTOS 2019 & RETFound\_mae & RETFound\_mae\_natureCFP.pth & 1024 & 224 & 224 & cuda & 8 \\
DDR & RETFound\_mae & RETFound\_mae\_natureCFP.pth & 1024 & 224 & 224 & cuda & 16 \\
\bottomrule
\end{tabularx}
\end{table}

This cached-feature design reduces compute cost, keeps the adaptation space small, and isolates the research question of how last-layer uncertainty affects screening-oriented operating points.

\subsection{Last-layer heads}
The deterministic baseline is a softmax linear head trained on cached RETFound features. Temperature scaling is applied post hoc on the validation predictions and therefore changes probability calibration without changing the ranking-based confusion matrix at a fixed threshold \citep{guo2017calibration}.

The variational Bayesian head models the final linear layer weights with diagonal Gaussian posteriors and optimizes cross-entropy plus a KL term. The Laplace baseline first trains a deterministic final layer, then fits a diagonal Gaussian approximation around the final-layer optimum. The SNGP-style head applies an optional spectral-normalized projection, fixed random Fourier features, and a diagonal precision estimate over the random-feature activations. This is an SNGP-style cached-feature head, not full end-to-end SNGP RETFound.

\subsection{Calibration, thresholding, and selective referral}
Three evaluation modes are central to this study.
\begin{enumerate}[leftmargin=1.2em]
\item \textbf{Full coverage}: every example receives an automated prediction.
\item \textbf{Threshold sweeps}: the positive-class probability threshold is varied to test whether low-false-negative behavior can be reached by ordinary threshold control.
\item \textbf{Selective referral}: the most uncertain examples are deferred to human review, and metrics are computed on accepted automated cases only.
\end{enumerate}

For selective referral, \emph{coverage} denotes the accepted fraction and \emph{referral rate} denotes the deferred fraction. This distinction matters because accepted-case performance should not be interpreted as full-population performance.

\subsection{Evaluation metrics}
We report accuracy, AUC, sensitivity, specificity, balanced accuracy, calibration metrics (ECE, NLL, Brier score), and confusion-matrix-derived false-positive and false-negative counts. The principal safety interpretation uses sensitivity, false negatives, and selective-referral behavior rather than accuracy alone.

\section{Experiments}

\subsection{APTOS internal validation}
APTOS is used to establish internal full-coverage baselines, threshold-sweep controls, and selective-referral behavior. The central question is whether uncertainty-aware last-layer heads can move the operating point toward fewer missed referable cases without collapsing specificity.

\subsection{DDR second-dataset validation}
DDR provides a harder second-dataset validation setting. Native DDR softmax and Bayesian heads test whether the qualitative APTOS findings reproduce when the frozen RETFound features are reused on a different retinal dataset with its own train/validation/test split.

\subsection{APTOS-to-DDR SNGP transfer evaluation}
The APTOS-trained SNGP sensitivity-selected checkpoint is evaluated directly on DDR cached features without retraining. The fitted diagonal precision state is restored from the APTOS checkpoint rather than recomputed on DDR. This makes the experiment a transfer test rather than a native DDR re-fit.

\section{Results}

\subsection{Full-coverage model results}
\begin{table}[t]
\centering
\small
\caption{APTOS validation comparison for full-coverage cached-feature heads.}
\label{tab:aptos-models}
\begin{tabular}{lrrrrrrrr}
\toprule
Model & Acc & AUC & Sens & Spec & Bal Acc & ECE & FN & FP \\
\midrule
Softmax & 0.8798 & 0.9580 & 0.8896 & 0.8726 & 0.8811 & 0.0348 & 17 & 27 \\
Softmax+Temp & 0.8798 & 0.9580 & 0.8896 & 0.8726 & 0.8811 & 0.0186 & 17 & 27 \\
Bayes (val-loss) & 0.8934 & 0.9650 & 0.8831 & 0.9009 & 0.8920 & 0.0216 & 18 & 21 \\
Bayes (max-sens) & 0.8934 & 0.9613 & 0.9610 & 0.8443 & 0.9027 & 0.0320 & 6 & 33 \\
Bayes (balanced) & 0.9016 & 0.9617 & 0.9545 & 0.8632 & 0.9089 & 0.0275 & 7 & 29 \\
Laplace (val-loss) & 0.8962 & 0.9666 & 0.8636 & 0.9198 & 0.8917 & 0.0972 & 21 & 17 \\
Laplace (sens) & 0.8962 & 0.9586 & 0.9481 & 0.8585 & 0.9033 & 0.0900 & 8 & 30 \\
SNGP (val-loss) & 0.9071 & 0.9756 & 0.8831 & 0.9245 & 0.9038 & 0.0288 & 18 & 16 \\
SNGP (sens) & 0.9098 & 0.9711 & 0.9805 & 0.8585 & 0.9195 & 0.0287 & 3 & 30 \\
\bottomrule
\end{tabular}
\end{table}

\begin{table}[t]
\centering
\small
\caption{DDR validation comparison for full-coverage native DDR heads and APTOS-to-DDR SNGP transfer.}
\label{tab:ddr-models}
\begin{tabular}{lrrrrrrrr}
\toprule
Model & Acc & AUC & Sens & Spec & Bal Acc & ECE & FN & FP \\
\midrule
DDR Softmax & 0.7939 & 0.8688 & 0.7773 & 0.8075 & 0.7924 & 0.0223 & 188 & 199 \\
DDR Bayes (val-loss) & 0.7966 & 0.8782 & 0.7275 & 0.8530 & 0.7902 & 0.0174 & 230 & 152 \\
DDR Bayes (sens) & 0.7678 & 0.8634 & 0.8472 & 0.7031 & 0.7751 & 0.0097 & 129 & 307 \\
APTOS->DDR SNGP & 0.5990 & 0.5833 & 0.2701 & 0.8675 & 0.5688 & 0.2168 & 616 & 137 \\
\bottomrule
\end{tabular}
\end{table}

APTOS full-coverage results show that several uncertainty-aware heads improved screening-oriented operating points relative to the cached softmax baseline. The sensitivity-selected variational Bayesian model reduced false negatives from 17 to 6. The sensitivity-selected Laplace model reduced false negatives to 8. The SNGP sensitivity-selected model reduced false negatives to 3 while keeping AUC at 0.9711 and balanced accuracy at 0.9195.

\begin{figure}[!htbp]
\centering
\includegraphics[width=0.82\textwidth]{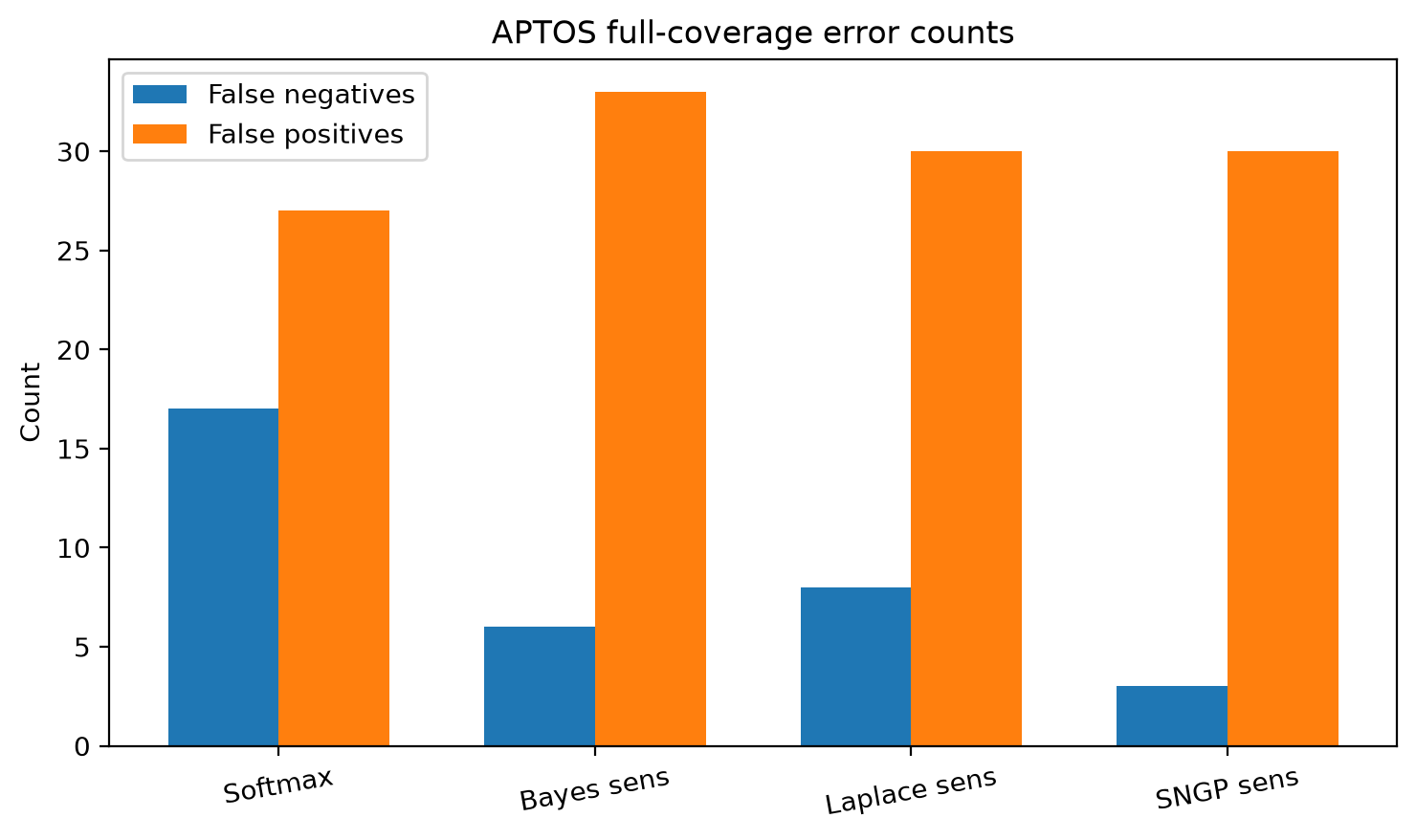}
\caption{APTOS full-coverage false-negative and false-positive counts for key internal models.}
\label{fig:aptos-full-coverage}
\end{figure}

These APTOS improvements did not justify a broad superiority claim. The softmax baseline plus temperature scaling remained a strong calibration reference, and the operating points differed by purpose. For example, the val-loss-selected Bayesian model improved NLL and Brier score relative to the deterministic baseline, while the sensitivity-selected Bayesian operating point emphasized false-negative reduction at a specificity cost.

DDR results were more sobering. The native DDR Bayesian sensitivity-selected model reduced false negatives from 188 to 129 relative to DDR softmax, which qualitatively reproduced the APTOS direction. However, specificity fell to 0.7031. The APTOS-to-DDR SNGP transfer result was a clear negative result: sensitivity fell to 0.2701, AUC to 0.5833, and false negatives rose to 616.

\begin{figure}[!htbp]
\centering
\includegraphics[width=0.84\textwidth]{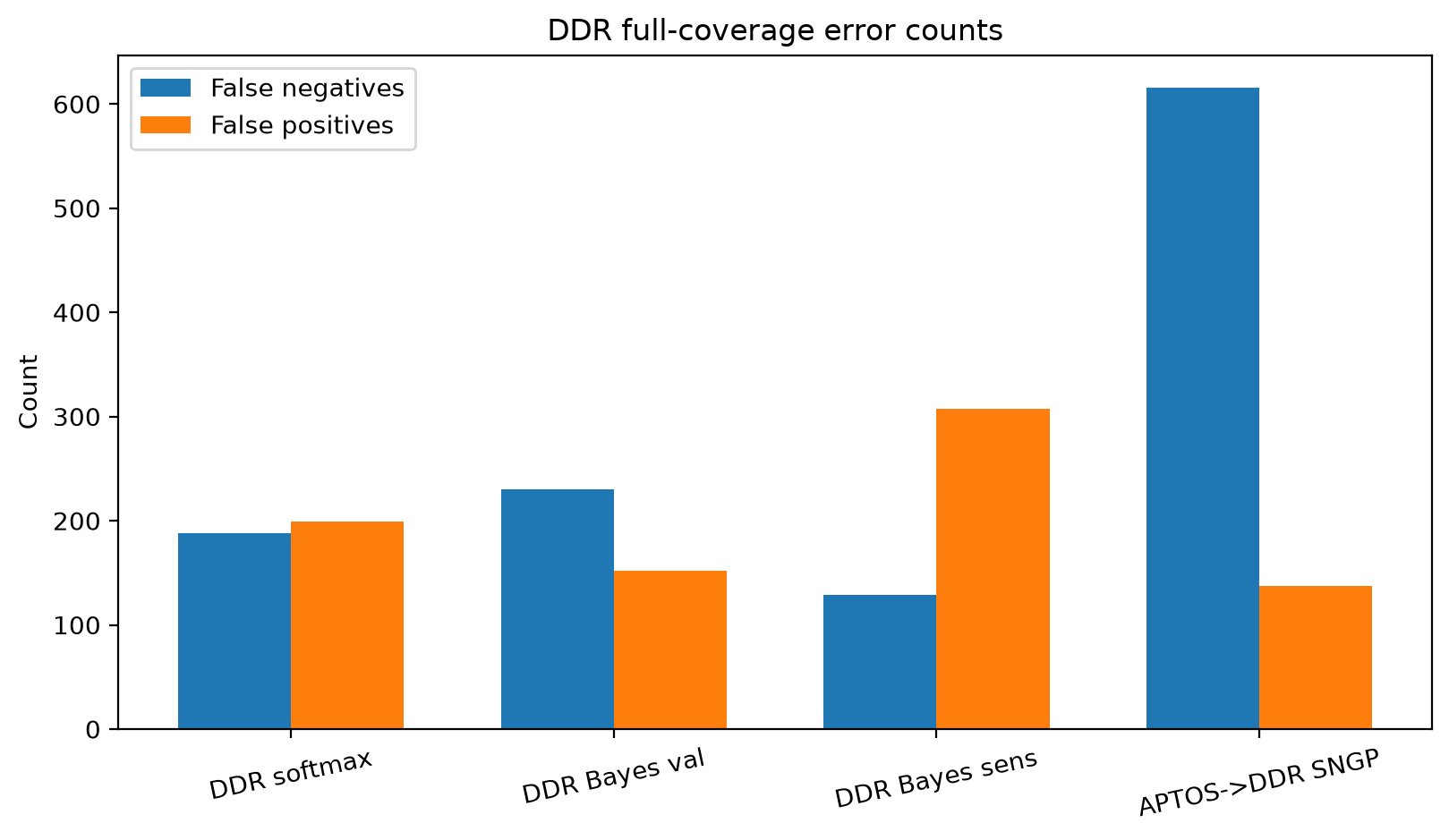}
\caption{DDR full-coverage false-negative and false-positive counts. The APTOS-trained SNGP transfer checkpoint performs poorly under shift.}
\label{fig:ddr-full-coverage}
\end{figure}

\begin{figure}[!htbp]
\centering
\includegraphics[width=0.82\textwidth]{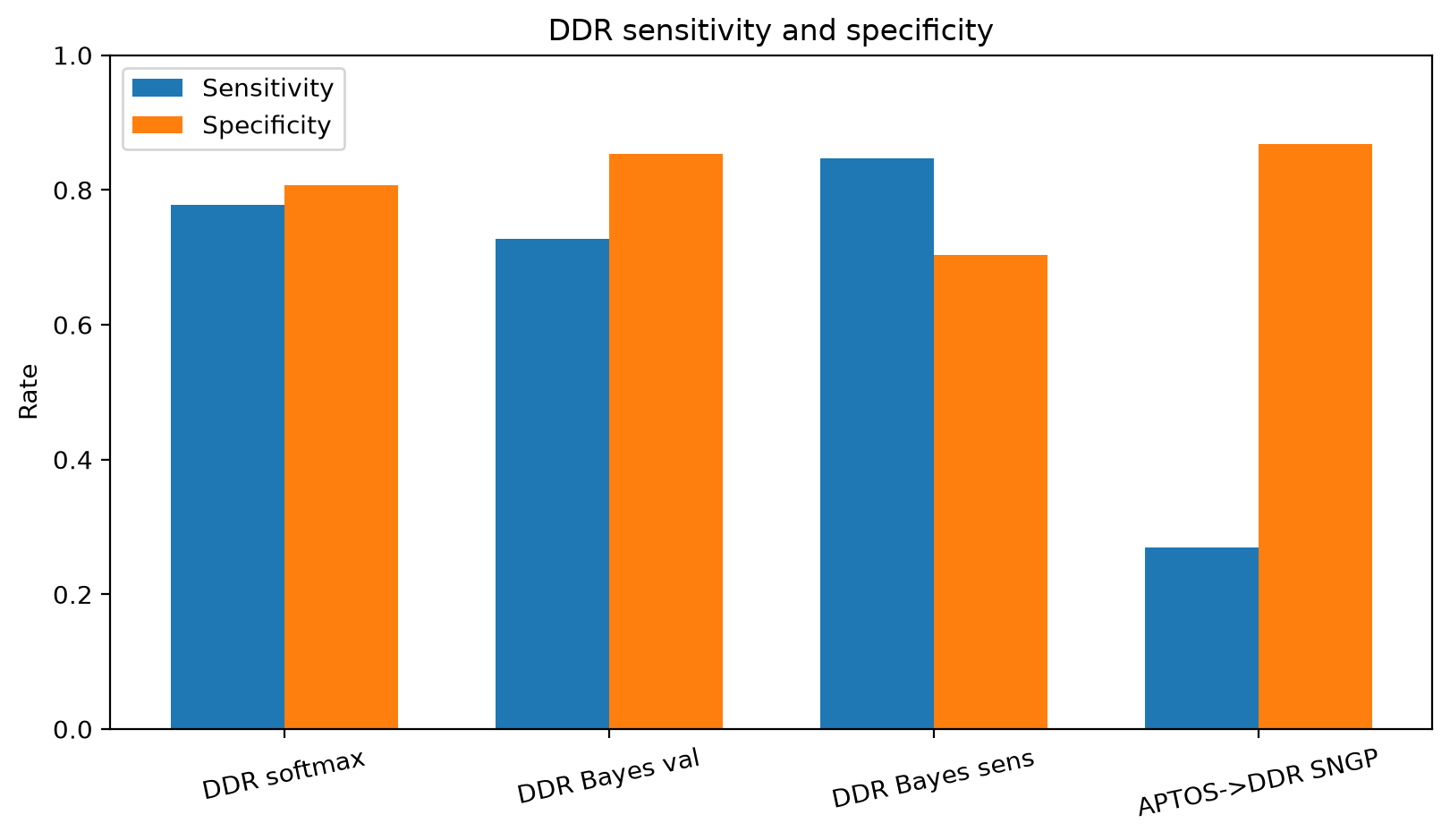}
\caption{DDR sensitivity and specificity comparison for native DDR heads and the APTOS-to-DDR SNGP transfer evaluation.}
\label{fig:ddr-sens-spec}
\end{figure}

\subsection{Threshold-sweep results}
\begin{table}[t]
\centering
\small
\caption{DDR threshold-sweep summary for balanced and low-false-negative operating points.}
\label{tab:ddr-thresholds}
\begin{tabular}{lrrrrrrr}
\toprule
Sweep & Policy & Thr & Sens & Spec & Bal Acc & FN & FP \\
\midrule
Softmax+Temp & Best bal acc & 0.50 & 0.7773 & 0.8075 & 0.7924 & 188 & 199 \\
Softmax+Temp & Lowest FN & 0.03 & 1 & 0.1025 & 0.5512 & 0 & 928 \\
Bayes (val-loss) & Best bal acc & 0.40 & 0.8104 & 0.7785 & 0.7945 & 160 & 229 \\
Bayes (sens) & Best bal acc & 0.55 & 0.8152 & 0.7524 & 0.7838 & 156 & 256 \\
Bayes (sens) & Lowest FN & 0.03 & 1 & 0.0532 & 0.5266 & 0 & 979 \\
APTOS->DDR SNGP & Best bal acc & 0.45 & 0.3069 & 0.8424 & 0.5746 & 585 & 163 \\
APTOS->DDR SNGP & Lowest FN & 0.01 & 0.9882 & 0.0155 & 0.5018 & 10 & 1018 \\
\bottomrule
\end{tabular}
\end{table}

Threshold sweeps showed that false-negative reduction was not unique to Bayesian modeling. On APTOS, the SNGP sensitivity-selected checkpoint reached zero false negatives at threshold 0.20 but required 67 false positives. The earlier cached softmax and Bayesian APTOS sweeps showed the same pattern: the softmax baseline, Bayesian max-sensitivity model, and Bayesian balanced model all reached zero false negatives at different low thresholds but converged to the same high-false-positive operating point.

On DDR, the same control remained important. The native DDR Bayesian sensitivity-selected model reached zero false negatives at threshold 0.03, but produced 979 false positives. The DDR temperature-scaled softmax model also reached zero false negatives at threshold 0.03, but with 928 false positives and specificity of only 0.1025. This is clinically different from selective referral because it treats almost the entire validation set as positive.

\begin{figure}[!htbp]
\centering
\includegraphics[width=0.82\textwidth]{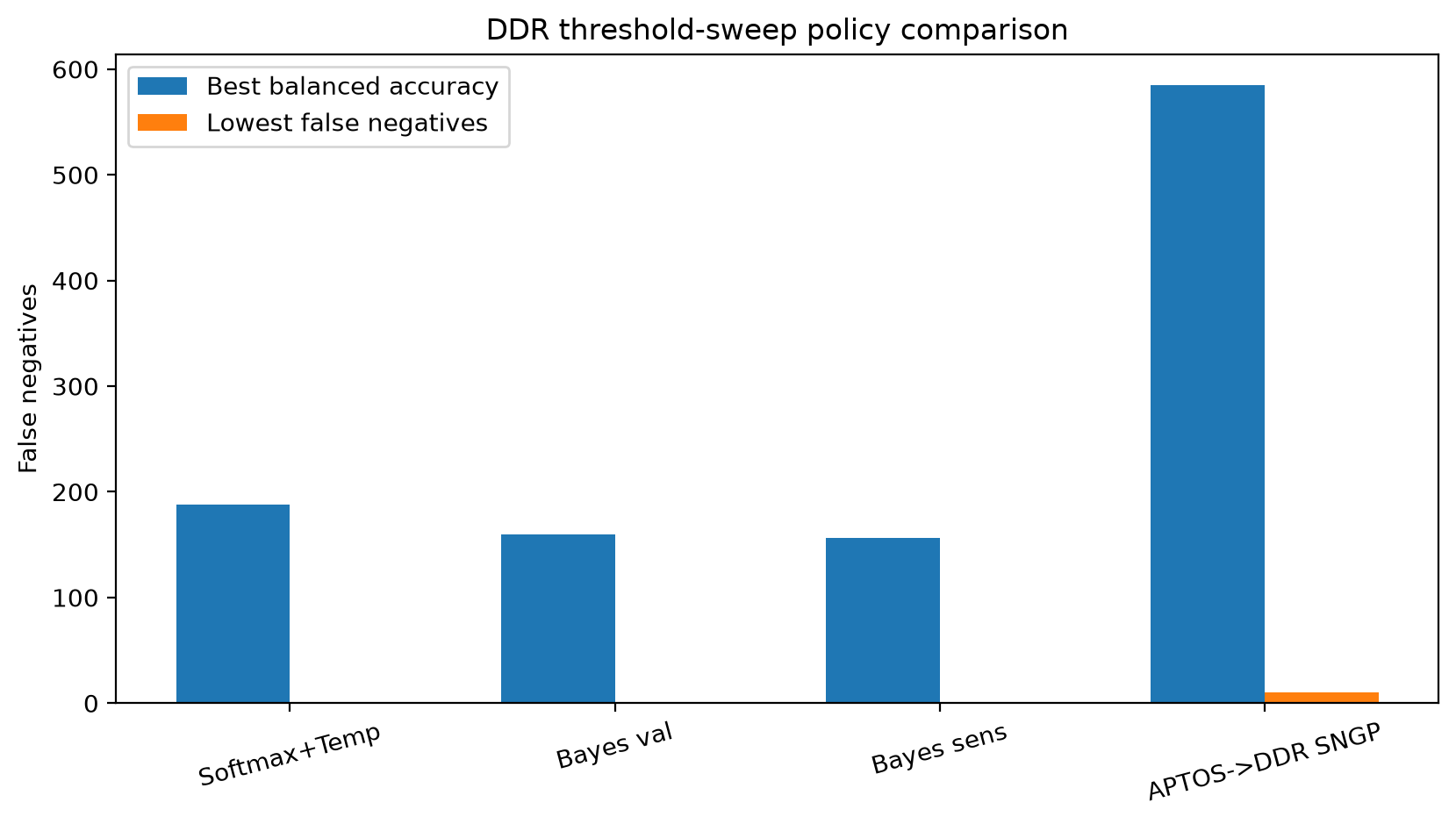}
\caption{DDR threshold-sweep comparison across balanced and low-false-negative operating points.}
\label{fig:ddr-thresholds}
\end{figure}

\subsection{Selective-referral results}
\begin{table}[t]
\centering
\footnotesize
\caption{DDR selective-referral comparison at approximately 80\% accepted coverage.}
\label{tab:ddr-selective}
\begin{tabularx}{\textwidth}{X r r r r r r r}
\toprule
Signal & Coverage & Referral & Sens & Spec & Bal Acc & FN & FP \\
\midrule
DDR Bayes confidence & 0.8003 & 0.1997 & 0.8966 & 0.7596 & 0.8281 & 72 & 194 \\
DDR Bayes entropy & 0.8003 & 0.1997 & 0.8966 & 0.7596 & 0.8281 & 72 & 194 \\
DDR Bayes prob var & 0.8003 & 0.1997 & 0.8887 & 0.7298 & 0.8093 & 75 & 224 \\
DDR Bayes MI & 0.8003 & 0.1997 & 0.8783 & 0.7284 & 0.8033 & 83 & 223 \\
APTOS->DDR SNGP entropy & 0.8003 & 0.1997 & 0.2160 & 0.9298 & 0.5729 & 519 & 59 \\
APTOS->DDR SNGP variance & 0.8003 & 0.1997 & 0.3042 & 0.8370 & 0.5706 & 478 & 133 \\
APTOS->DDR SNGP combined & 0.8003 & 0.1997 & 0.2165 & 0.9284 & 0.5725 & 521 & 60 \\
\bottomrule
\end{tabularx}
\end{table}

Selective referral was the strongest internal APTOS safety result. For the SNGP sensitivity-selected checkpoint, predictive entropy at approximately 80\% coverage deferred 73 of 366 cases and reduced accepted-case false negatives to zero while preserving accepted-case specificity of 0.9146 and accepted-case accuracy of 0.9522.

On DDR, native Bayesian uncertainty signals remained useful but weaker. At approximately 80\% coverage, DDR Bayesian confidence or predictive entropy reduced accepted-case false negatives from 129 to 72, while improving accepted-case balanced accuracy to 0.8281. Probability variance and mutual information were weaker than confidence or entropy on DDR.

By contrast, the APTOS-trained SNGP transfer checkpoint did not yield useful DDR selective-referral behavior. At approximately 80\% coverage, its entropy-based referral still left 519 accepted-case false negatives, and the variance-only signal left 478. The problem was not simply poor calibration; the cross-dataset ranking itself was weak.

\begin{figure}[!htbp]
\centering
\includegraphics[width=0.84\textwidth]{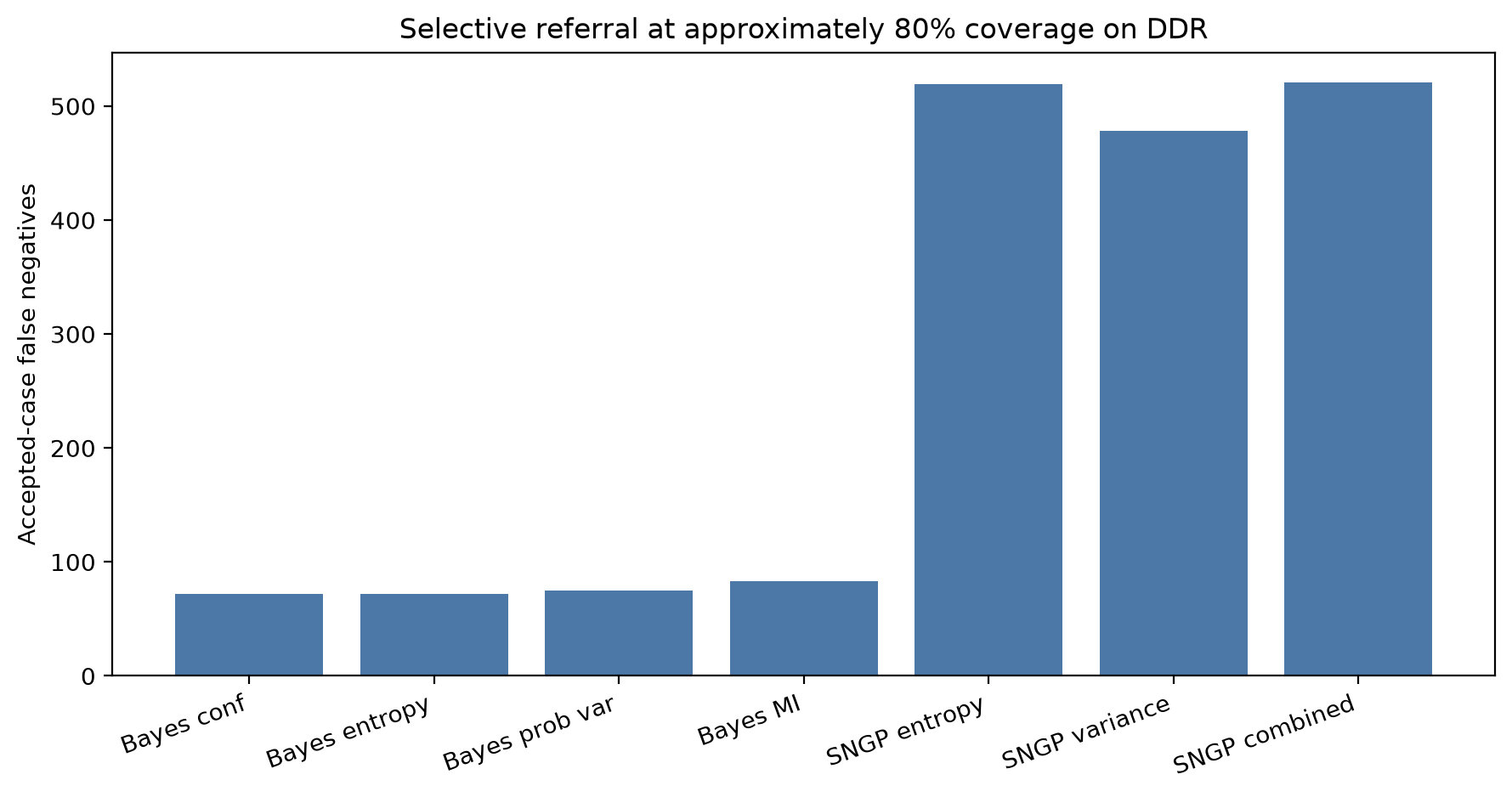}
\caption{Accepted-case false negatives at approximately 80\% coverage for DDR Bayesian and APTOS-to-DDR SNGP uncertainty signals. Lower is better.}
\label{fig:ddr-selective}
\end{figure}

\subsection{SNGP transfer failure}
\begin{table}[t]
\centering
\small
\caption{SNGP internal APTOS validation versus APTOS-to-DDR transfer evaluation.}
\label{tab:sngp-transfer}
\begin{tabular}{lrrrrrrrr}
\toprule
Setting & Acc & AUC & Sens & Spec & Bal Acc & ECE & FN & FP \\
\midrule
APTOS internal & 0.9098 & 0.9711 & 0.9805 & 0.8585 & 0.9195 & 0.0287 & 3 & 30 \\
APTOS->DDR transfer & 0.5990 & 0.5833 & 0.2701 & 0.8675 & 0.5688 & 0.2168 & 616 & 137 \\
\bottomrule
\end{tabular}
\end{table}

\begin{figure}[!htbp]
\centering
\includegraphics[width=0.78\textwidth]{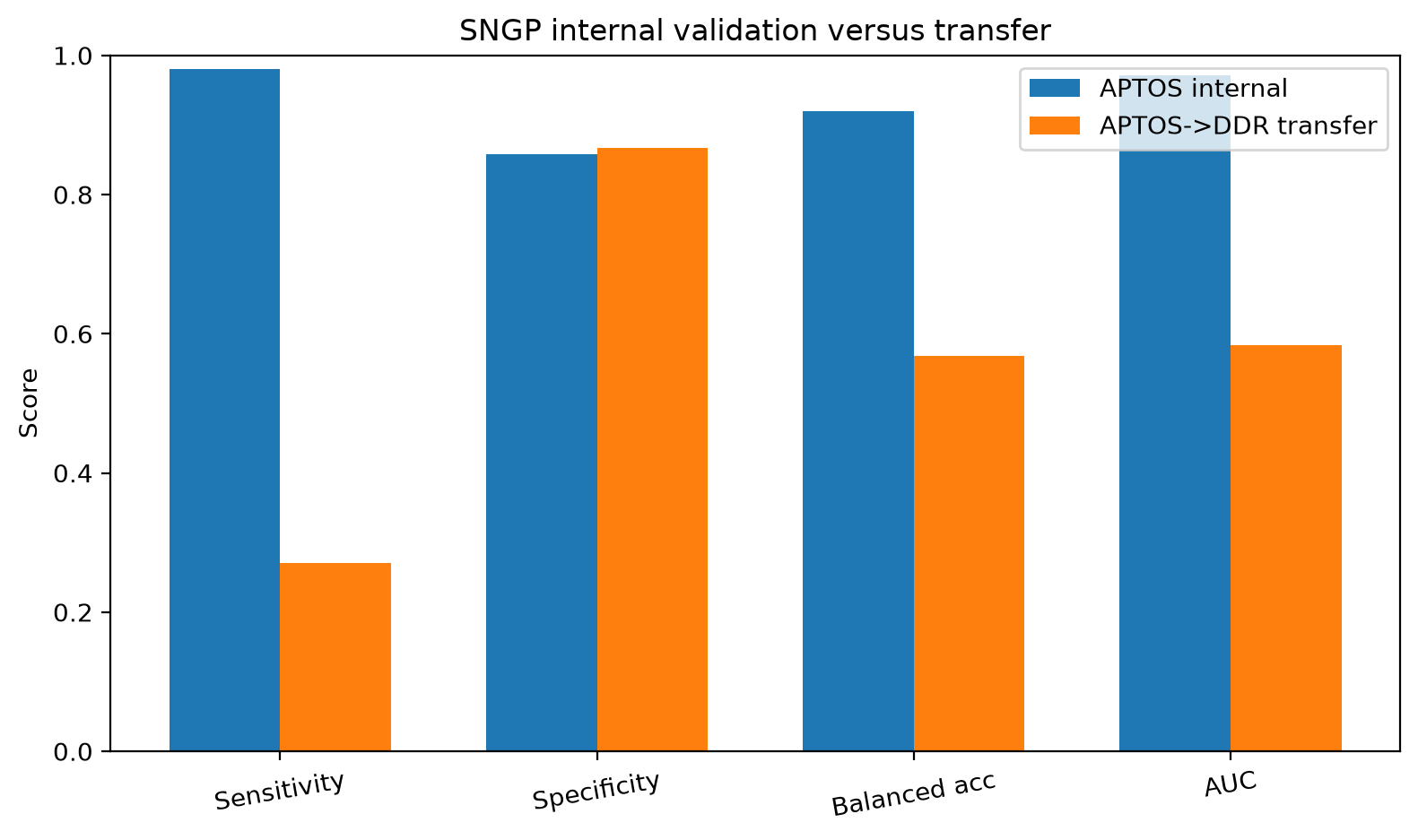}
\caption{APTOS internal validation versus APTOS-to-DDR transfer for the sensitivity-selected SNGP checkpoint.}
\label{fig:sngp-transfer}
\end{figure}

The SNGP internal-versus-transfer comparison is the clearest negative result in the study. Internally on APTOS, the sensitivity-selected SNGP model achieved sensitivity 0.9805, balanced accuracy 0.9195, and only 3 false negatives. The same checkpoint evaluated directly on DDR without retraining dropped to sensitivity 0.2701 and balanced accuracy 0.5688.

This should be interpreted conservatively. The result does not show that SNGP is universally ineffective; it shows that this SNGP-style cached-feature implementation did not improve cross-dataset robustness in this experiment.

\FloatBarrier
\section{Discussion}

\subsection{What worked}
Three findings were consistently useful.
\begin{enumerate}[leftmargin=1.2em]
\item Frozen RETFound cached features were strong enough to support informative last-layer comparisons on both datasets.
\item Variational Bayesian last-layer heads improved screening-oriented full-coverage operating points on both APTOS and DDR relative to the default deterministic baseline.
\item Selective referral meaningfully improved accepted-case safety when the uncertainty signal preserved a useful ranking, especially for APTOS and for the native DDR Bayesian sensitivity-selected model.
\end{enumerate}

\subsection{What did not transfer}
The APTOS-trained SNGP checkpoint did not transfer well to DDR. Its full-coverage discrimination was weak, its calibration was poor, and its uncertainty signals did not reliably surface dangerous missed-positive cases under shift. This is an important negative result because internal validation alone could have suggested a stronger robustness story than the data support.

\subsection{Why threshold tuning matters}
Threshold tuning is a necessary control because it limits overclaiming. If a deterministic or Bayesian model can eliminate false negatives simply by lowering the decision threshold, then zero false negatives alone cannot support a claim that Bayesian uncertainty was uniquely responsible. The relevant comparison is the full tradeoff among false negatives, false positives, specificity, and human-review burden.

\subsection{Why selective referral is clinically meaningful}
Selective referral addresses a different workflow from threshold-only screening. Lowering the threshold pushes more cases into automated positive predictions. Selective referral instead defers uncertain cases to human review. In this study, that workflow preserved much stronger accepted-case accuracy and specificity than low-threshold zero-false-negative screening on APTOS, and it remained meaningfully useful on DDR for the native Bayesian model.

\subsection{Limitations}
This is not a clinical deployment study. It is a research-grade empirical evaluation on two retinal datasets using frozen RETFound features. The study is limited by the absence of raw-image fine-tuning, limited uncertainty families, and evaluation on a single external dataset rather than multiple independent cohorts. The manuscript package in this repository also reconstructs publication tables from locked summary results because the full raw experiment output directories are not included in this checkout.

\section{Conclusion}
Uncertainty-aware last-layer adaptation on frozen RETFound features produced useful internal safety-oriented operating points, especially when evaluated through selective referral rather than accuracy alone. However, the study also shows three constraints on what can be honestly claimed. First, false-negative reduction is not unique to Bayesian modeling because threshold tuning can also reduce false negatives at substantial false-positive cost. Second, uncertainty signals that look useful internally may weaken or fail under external dataset shift. Third, negative transfer results such as the APTOS-to-DDR SNGP evaluation are essential evidence, not side notes. The practical conclusion is that uncertainty-aware retinal screening systems should be evaluated using explicit safety--coverage tradeoffs and second-dataset validation, rather than relying on aggregate accuracy or in-distribution gains alone.

\section{Reproducibility}
\begin{table}[t]
\centering
\footnotesize
\caption{Reproducibility metadata for the manuscript package.}
\label{tab:reproducibility}
\begin{tabularx}{\textwidth}{p{0.23\textwidth}X}
\toprule
Item & Value \\
\midrule
Repository & \url{https://github.com/kmardhani/uncertainty-aware-retfound} \\
Author & Karim Mardhani \\
Affiliation & Master of Science in Artificial Intelligence; University of Colorado Boulder; Boulder, CO, USA \\
Email & \nolinkurl{Karim.Mardhani@colorado.edu} \\
Title & Uncertainty-Aware Last-Layer Adaptation of RETFound for Referable Diabetic Retinopathy Screening Under Dataset Shift \\
arXiv primary & cs.CV \\
arXiv secondary & cs.LG \\
Project status & Independent research initiative; not sponsored, supervised, or formally endorsed by the university \\
Task & Binary referable diabetic retinopathy \\
Backbone & Frozen RETFound\_mae \\
Feature dimension & 1024 \\
APTOS split & 2930 / 366 / 366 \\
DDR split & 8765 / 1878 / 1879 \\
\bottomrule
\end{tabularx}
\end{table}

The repository contains scripts for metadata preparation, frozen feature extraction, cached-feature training, calibration, threshold sweeps, selective referral, and manuscript asset generation. The paper assets can be regenerated from the repository root with:
\begin{verbatim}
uv run python paper/scripts/make_paper_assets.py
\end{verbatim}
A stable release tag should be used for archival reproduction.

\section*{Acknowledgments}
This work was conducted as an independent research initiative by the author. The author is enrolled in the part-time, course-based Master of Science in Artificial Intelligence program at the University of Colorado Boulder; however, this project was not sponsored, supervised, or formally endorsed by the university.

\appendix
\section{Additional Notes on Clinical Interpretation}
Accepted-case selective-referral metrics are conditional on non-deferred cases and should not be interpreted as full-population screening performance. Likewise, the APTOS-to-DDR SNGP result should be interpreted specifically as a transfer evaluation without retraining, not as a statement about every possible SNGP design.

\bibliographystyle{plainnat}
\bibliography{references}

\end{document}